# Defocused images' removal of axial overlapping scattering particles by using three-dimensional nonlinear diffusion based on digital holography

**WEI-NA LI,[1] ZHENGYUN ZHANG,[2] PING SU,[1,*] JIANSHE MA,[1] AND XIAOHAO WANG[1]**

[1]*Graduate School at Shenzhen, Tsinghua University, Shenzhen, Guangdong 518055, China*
[2]*BioSyM IRG, Singapore-MIT Alliance for Research and Technology (SMART) Centre, 1 CREATE Way, #04-13/14 Enterprise Wing, Singapore 138602, Singapore*
*\*su.ping@sz.tsinghua.edu.cn*

**Abstract:** We propose a three-dimensional nonlinear diffusion method to implement the similar autofocusing function of multiple micro-objects and simultaneously remove the defocused images, which can distinguish the locations of certain sized scattering particles that are overlapping along z-axis. It is applied to all of the reconstruction slices that are generated from the captured hologram after each back propagation. For certain small sized particles, the maxima of maximum gradient magnitude of each reconstruction slice appears at the ground truth *z* position after applying the proposed scheme when the reconstruction range along *z*-axis is sufficiently long and the reconstruction depth spacing is sufficiently fine. Therefore, the reconstructed image at ground truth *z* position is remained, while the defocused images are diffused out. The results demonstrated that the proposed scheme can diffuse out the defocused images which are 20 μm away from the ground truth *z* position in spite of that several scattering particles with different diameters are completely overlapping along *z*-axis with a distance of 800 μm when the hologram pixel pitch is 2 μm. It also demonstrated that the sparsity distribution of the ground truth z slice cannot be affected by the sparsity distribution of corresponding defocused images when the diameter of the particle is not more than 35um and the reconstruction depth spacing is not less than 20 μm.

## 1. Introduction

Unlike traditional microscopy that merely focuses on one plane, digital holography can capture a volume, and reconstruct every plane of this volume. Nowadays, the sizing, counting, and locating issues of *in situ* micro-objects (bubbles, particles, or microorganisms) gain a wide variety of attentions, especially when in-line digital holography is developed and becomes a promising microscopic alternative [1-2]. The four papers [3-6] employed spherical wave as the reference wave to illuminate opaque particles, rod-shaped particles, and transparent particles suspended in water. The authors located the positions and measured the size of each particle from the captured hologram. Its experimental configuration is very compact, and provides a magnification on the target objects. However, it reduces the field of view (FOV), since the traditional Fresnel diffraction is utilized, the accuracy of particle amount and the location along *z*-axis were significantly affected by the chosen depth spacing. Tian *et al.* employed plane wave as reference beam to illuminate bubbles [7], wherein the minimum intensity is used as a focus metric to detect edges of the bubbles, thereby locating each bubble's location, especially axial position. Despite it maintained large FOV and the processing time is faster than traditional Fresnel transform, the location is not accurate, especially when the bubble size is less than 10 μm, Moreover, when several bubbles is clustering together, this scheme recognizes them as one bubble. Overall, the traditional diffraction-based method can cause severe defocused-image issue under the condition that the ground truth $z$ position of each micro-object is unknown. It directly leads to the inaccuracy of micro-objects on the mount and locations.

Autofocusing is considered as one means to obtain the exact location of the object in digital holography (DH) and digital holographic microscopy (DHM). On one hand, experimental configuration setup can implement autofocusing [8-9]. On the other hand, autofocusing is mostly implemented by computation algorithms, for instance, image sharpness [10-11], structure tensor [12], edge sparsity [13], and magnitude differential [14]. All of these algorithms are merely suitable for large single object or a microscopic plane sample, and the reconstructed image of each investigated distance should be obtained during the focusing distance searching process. Subsequently, a focus metric is applied to achieve a series of values corresponding to the series of reconstructed images, eventually, a peak value or a valley value denotes the focused distance. However, Ren *et al.* [15] took advantage of convolutional neural network (CNN) to implement autofocusing as a classification problem, and provide rough estimates of the focusing distance with each classification. It is still more appropriate for large single object, even though reconstructed images are not requested.

Constructing a compressive model with sparsity shows a good performance when encountering noise and ghost images by converting hologram reconstruction problem to a regularized nonlinear optimization. Brady *et al.* introduced compressive sensing algorithm into digital holography, they demonstrated that decompressive inference can infer multidimensional objects from a two-dimensional (2D) hologram [16]. Liu *et al.* applied compressive holography to object localization [17-18], which has shown orders of magnitude improvement on lateral localization accuracy so long as the solution is sparse in its derivative. Chen *et al.* used plane wave to illuminate bubbles [19], and they suggested the compressive holographic method to locate the axial position of each bubble. It cannot distinguish the bubbles which were completely or partially overlapping along z-axis. In addition, if the amount of bubbles is higher than 256 with the reconstruction depth spacing of 250 μm, the results were as same as the traditional Fresnel transform. A phase iteratively enhanced compressive sensing (PIE-CS) algorithm can achieve phase imaging and eliminate defocused images simultaneously [20]. Total variation (TV) regularizer was employed in all of the aforementioned papers. Tian *et al.* [21] suggested a two-dimensional hybrid-Weickert nonlinear diffusion (2D HWNLD) scheme in the research of transport of intensity equation (TIE) to remove the low frequency artifacts on an image effectively. Zhang *et al.* [22] proposed a method including both simultaneous measurement and reconstruction tailoring for quantitative phase imaging. A mild sparsity promoter regularizer is utilized to minimize the expected end-to-end error and to yield optimal design parameters for both the measurement and reconstruction processes of the thin phase object.

We propose a three-dimensional nonlinear diffusion method to implement the similar autofocusing function of multiple micro-objects and simultaneously remove the defocused images, which can distinguish the locations of certain sized scattering particles that are overlapping along $z$-axis as well. It is applied to all of the reconstructed images that are generated from the captured hologram after each back propagation. For certain small sized particles, the maxima of the gradient magnitude curve of all of the reconstruction slices appears at the ground truth $z$ position after applying the proposed scheme when the reconstruction range along $z$-axis is sufficiently long and the reconstruction depth spacing is sufficiently fine, therefore, the focused image at the ground truth $z$ position is remained, however, the defocused images are diffused out. The numerical results demonstrated that the proposed scheme can diffuse out the defocused images which are 20 μm away from the ground truth $z$ position of the particle with the diameter of 15 μm and with the hologram pixel pitch of 2 μm, and demonstrated that completely $z$-axis overlapping particles can be resolved as well. To our knowledge, this is the first time that a three-dimensional nonlinear diffusion regularizer is utilized in digital holography to solve depth location issues, and it is found effective to distinguish the overlapping particles along z-axis, while preserve the advantage in the elimination of defocused images.

The remainder of the paper is organized as follows. Three-dimensional hybrid-Weickert nonlinear diffusion (3D HWNLD) regularizer is introduced in Section 2. In Section 3, the demonstrations of defocused image removal are presented, which includes single-particle case and overlapping-particle case. Finally, the conclusions are summarized in Section 4.

## 2. Three-dimensional hybrid-Weickert nonlinear diffusion regularizer

We investigated the impact of the 3D HWNLD scheme on the scattering particles suspended in the water. In the optical setup, a plane wave illuminates a cuvette in which plenty of particles suspended in the milli-Q water. Subsequently, charge-coupled device (CCD) camera captures the hologram of the whole volume. The schematics is shown below in Fig. (1). Assume there are $m$ particles in the captured volume $V$, and each particle $particle_i(\xi,\eta)$ corresponds to a distance $z_i$ away from CCD, whose Fresnel transform $H_i(x,y,z_i)$ is shown in Eq.(1) [23], and the summation $H(x,y)$ of the Fresnel transforms of all the $m$ particles is shown in Eq. (2). Since the plane wave is utilized, assume its intensity is 1 for the simplification. Hence the mathematical expression of the final hologram $I$ captured by CCD is shown in Eq. (3), in which the plane wave is omitted.

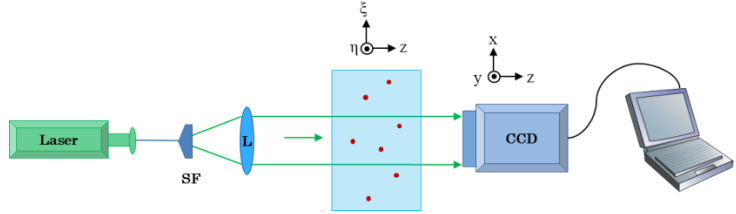

Fig. 1. Schematics of the optical setup, SF: spatial filter, L: collimating lens.

$$H_i(x,y,z_i) = FT^{-1}\begin{Bmatrix} \exp(jkz_i) \times FT\{particle_i(\xi,\eta)\} \\ \times \exp(-j\pi\lambda z_i(f_x^2 + f_y^2)) \end{Bmatrix}, \quad (1)$$

$$H(x,y) = \sum_i^m H_i(x,y,z_i) \quad (2)$$

$$I(x,y) = 1 + H^2(x,y) + H(x,y) + H^*(x,y) + n, \quad (3)$$

where $(\xi,\eta)$, $(f_x,f_y)$, and $(x,y)$ represent the lateral coordinates of the particle's position, spatial frequency domain, and hologram plane, respectively, $n$ denotes noise. $FT\{\cdot\}$ and $FT^{-1}\{\cdot\}$ denote Fourier transform and inverse Fourier transform, respectively [14]. $H^*(x,y)$ stands for the complex conjugate of $H(x,y)$. In this application, reconstruction performance of $H(x,y)$ is rarely affected by $H^2(x,y)$ and $H^*(x,y)$.

It is difficult to obtain the exact $z$ position of the object from one hologram especially while the objects are highly concentrated (several hundred in one $cm^3$) with small size (in μm), which is caused by the limitation of the axial resolution of the hologram. The lateral resolution and axial resolution of the hologram are $\Delta x = \Delta y = \frac{\lambda z}{2d}$ and $\Delta z = \frac{2\lambda z^2}{d^2}$, respectively, where $d$ is the height of the camera chip and $z$ is the distance to camera. Hence, 3D HWNLD regularizer is introduced to defuse out the defocused images and remain the in-focus images in the reconstruction stack of a hologram. The objective function is shown in Eq. (4).

$$obj\_fun\ (x_{re}) = minimize\left\{\frac{1}{2}\|y_{sum} - Ax_{re}\|^2 + \tau\iiint \Psi(|\nabla x_{re}|)dxdydz\right\}, \qquad (4)$$

where $y_{sum}$ denotes the captured hologram that is literally the $I(x,y)$ in Eq. (3) and $x_{re}$ denotes a stack of reconstructed images with certain depth spacing, respectively; $A$ represents the forward propagation that is the Fresnel transform shown in Eq. (1), but with a series of distances simultaneously; $\tau$ denotes an user-chosen regularization parameter, and $\Psi(|\nabla x_{re}|)$ represents the integral of the hybrid-Weickert function [21], in which $\nabla$ stands for gradient operation on $x$, $y$, $z$ axes, respectively; an additional partial derivative along the $z$-axis is utilized in order to obtain the precise localization along z-axis, $x_{re}$ is an estimate of $x_{re}$. It is noted that the partial derivative along $z$-axis must be multiplied by a constant $\frac{pixel\ pitch}{depth\ spacing}$ to scale it as same as the magnitude order of the partial derivatives of $x$ and $y$ axes. The final goal of 3D HWNLD is to obtain an extremely small change between two consecutive $obj\_fun\ (x_{re})$. The implementation of the 3D NLD is shown in Eq. (5).

$$\frac{\partial x_{re}}{\partial t} = \tau \nabla \cdot [FLX \cdot \nabla x_{re}], \qquad (5)$$

$$FLX = \frac{\Psi'(s)}{|\nabla x_{re}|}, \qquad (6)$$

where the flux function $FLX$ is composed of $\Psi'(s)$ and $\nabla x_{re}$, $\Psi'(s) = F_H(s)$ is the derivative of $\Psi(s)$, and $F_H(s)$ is the hybrid-Weickert function [21] shown in Eq. (7), in which 1.6s is removed from Weickert function $F_W = 1.6s[1 - \exp(-3.86/s^{12})]$ [24] for removing low frequency efficiently; $t$ is the time that is carried out by $\tau$ multiplied by *step-size* that is controlled by the iterative shrinkage/thresholding (IST) algorithm. Usually, the initialization of *step-size* is 1, its value exponentially decreases when the program runs.

$$\Psi'(s) = F_H(s) = 1 - \exp(\frac{-3.86}{s^{12}}), \qquad (7)$$

where $s = |\nabla x_{re}|/k0$, and $k0$ is a constant. If the gradient magnitude of the voxel which is higher than $k0$ is encountered, it is preserved (defused out slowly); otherwise, it is diffused out fast. However, compared with total variation (TV) diffusion regularizer, in which $\Psi'(x) = 1$ and $\Psi(x_{re}) = |\nabla x_{re}|$, HWNLD regularizer is more flexible, and the preserving ability is stronger. The curve illustrations of $F_H(s)$, flux equation of HWNLD with k0=1/3, and flux equation of TV are shown in Fig. 2. HWNLD is a convex curve like TV, therefore, it is a gradient decent convergence, eventually, IST program would stop when the first part balances the second part in the right-hand side of Eq. (4).

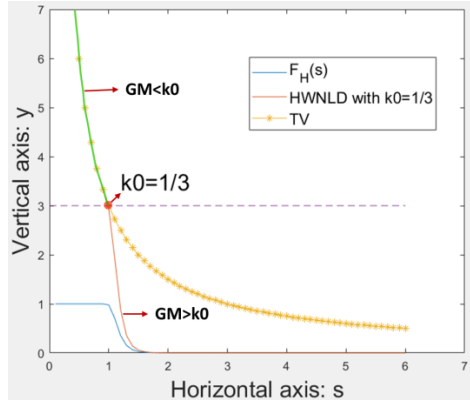

Fig. 2. Curve illustrations of $F_H(s)$, flux equation of HWNLD with $k_0=1/3$, and flux equation of TV, GM: gradient magnitude.

Before implementing 3D HWNLD, $\tau$ is settled, and $x_{re}$ is initialized by inverse Fresnel transform of the original hologram $I(x,y)$. Afterwards, 3D HWNLD is applied on $x_{re}$ to obtain a stack of estimated reconstructed images $x_{re}$, the corresponding $\Phi(|\nabla x_{re}|) = \iiint \Psi(|\nabla x_{re}|)dxdydz$ is calculated. If the minimization change $\Delta obj\_fun^{(i)}(x_{re})$, which is shown in Eq. (8) and the superscript $i$ represents the current iteration, is not less than a previously settled threshold, $x_{re}$ goes into next forward propagation to obtain a new hologram $y_{sum}$. Then, the back propagation is carried out on the residual hologram $(y_{sum} - y_{sum})$ to obtain a new stack of reconstructed images $resi\_x_{re}$ multiplied by a time $t$ ($t = \tau * step\_size$) and $x_{re}$ is replaced. Subsequently, apply 3D HWNLD on $x_{re}$ and replace $x_{re}$, then continue to carry out the loop until $\Delta obj\_fun^{(i)}(x_{re})$ is less than the settled threshold and end this program. This process is implemented by IST algorithm, and the flowchart of the main program framework is shown in Fig. 3.

$$\Delta obj\_fun^{(i)}(x_{re}) = \frac{\left|obj\_fun^{(i)}(x_{re}) - obj\_fun^{(i-1)}(x_{re})\right|}{obj\_fun^{(i-1)}(x_{re})}, \tag{8}$$

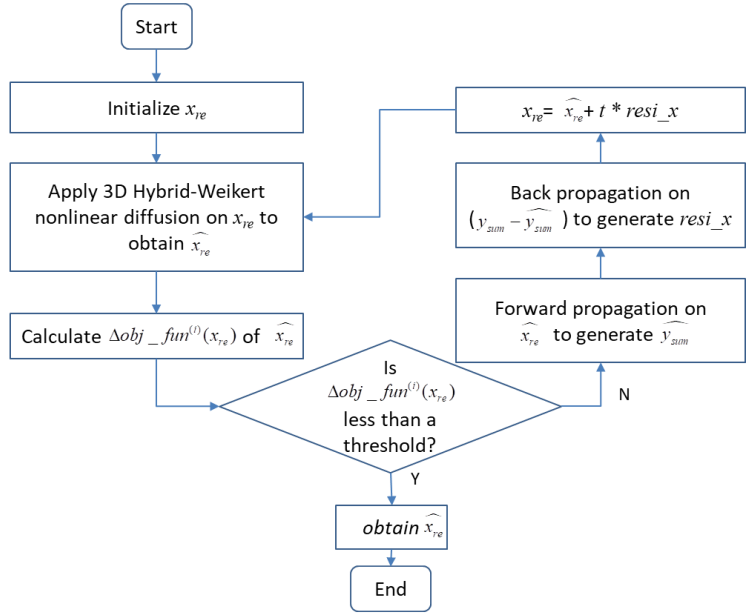

Fig. 3. Flowchart of the main program framework in IST.

## 3. The removal of defocused images

### 3.1 Scattering particles with different sizes

When a plane wave illuminates particles from water, several reflections and refractions occur before light leaving the rear surface of the particle. In this paper, the chosen physical model of the particle (refractive index = 1.58) is the Mie scattering particle with diameters from 10 μm to 60 μm, the centroid patterns of several particles are shown in Fig. 4. Several applications of Mie scattering model is employed in digital holographic microscopy [25-28]. It is noted that the edge is the lightest portion after subtracting the background noise, which implies the maximum gradient magnitude of the pixel is at the edge of the particle. Mie scattering model is a convenient and accurate way to generate simulated holograms, while the reconstruction doesn't depend on it.

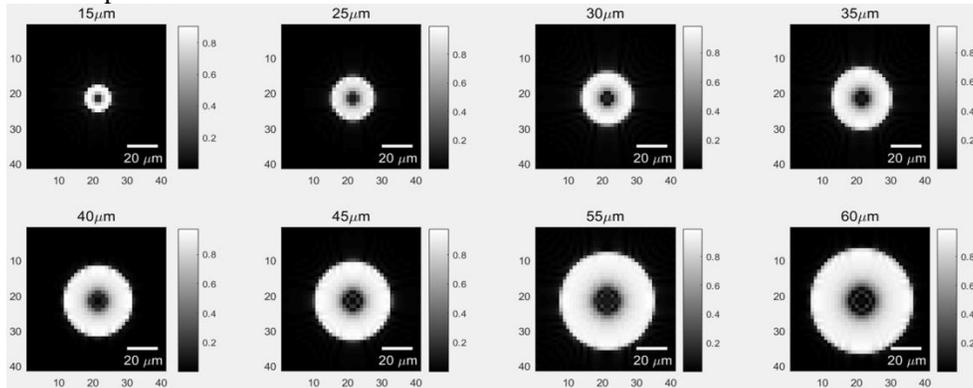

Fig. 4. Mie scattering particle models in the centroid position with different diameters, each image size: $41 \times 41$.

### 3.2 Single-particle case

In order to render to be similar to the real experiment, 1% Gaussian noise and 5000 photons per pixel Poisson noise are both added in all of the optical simulations. It is observed that the centroid pattern of the scattering particle is a ring pattern according to Fig. 4. Fig. 5 shows that the maximum intensity of each reconstruction slice in the original reconstruction stack (101 reconstructed slices along *z*-axis, namely, the range of reconstruction distance is [4338 μm, 6338 μm]) of each sized particle (diameters equal 15 μm, 25 μm, 30 μm, 35 μm, 40 μm, 45 μm, and 55 μm, respectively), and the depth spacing of reconstruction equals 20 μm between two neighboring reconstruction slices. The utilized wavelength is 660 nm, and pixel pith equals 2 μm in a 256×256 hologram. The original reconstruction stack is the all of reconstructed images generated from the hologram by utilizing the traditional Fresnel back propagation method. The intensity becomes higher and higher and reaches a peak at the two sides of the ground truth *z* position, moreover, the positions of the peak intensity appear further and further away from the ground truth *z* slice (GT) when the diameter becomes bigger and bigger. The hologram, whose mathematical expression is shown in Eq. (3), of scattering particle with the diameter of 15 μm is depicted in Fig. 6(a).

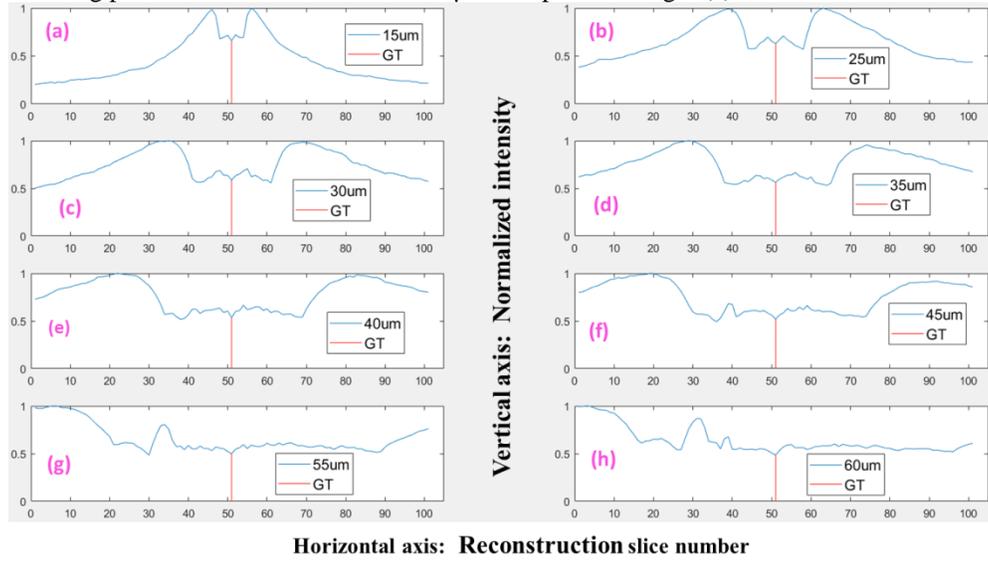

Fig. 5. The maximum intensity of each reconstruction slice in the reconstruction stack of single scattering particle hologram with diameter of (a) 15 μm, (b) 25 μm, (c) 30 μm, (d) 35 μm, (e) 40 μm, (f) 45 μm, (g) 55 μm, (h) 60 μm, GT: ground truth z slice. (depth spacing=20 μm, GT=51, at 5338 μm away from CCD)

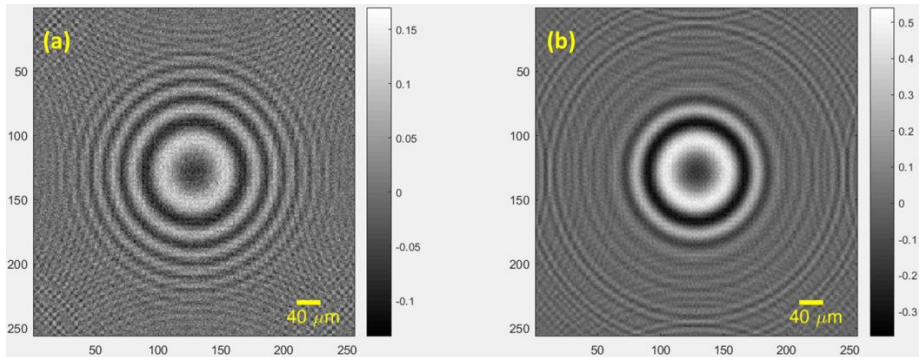

Fig. 6. (a) Hologram of one single particle (particle diameter=15 μm), (b) hologram of three overlapping particles along *z*-axis (particle diameter=20 μm, 15 μm, and 25 μm, respectively), hologram size: 256×256 .

It depicts the maximum gradient magnitude of each reconstruction slice in the original reconstruction stack of each sized particle's hologram in Fig. 7 wherein the slice number of GT is 51. The maxima of each curve does not always appear at the ground truth $z$ position, and the jamming of the defocused images along $z$-axis becomes so severe that the distribution of edge sparsity in the GT is affected when the particle size becomes bigger and bigger.

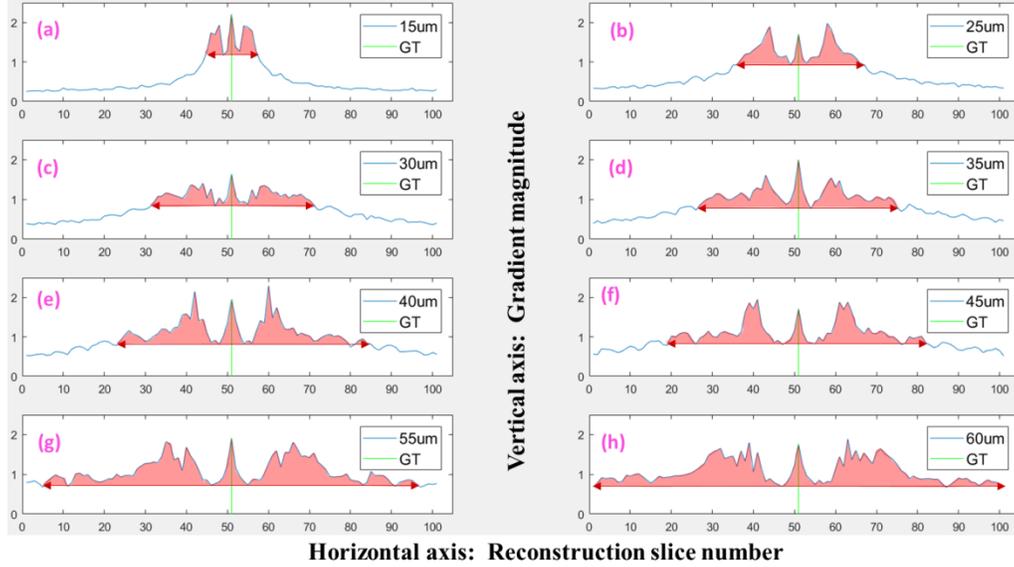

Fig. 7. The maximum gradient magnitude of each reconstruction slice in the reconstruction stack, the hologram of single scattering particle with the diameter of (a) 15 μm, (b) 25 μm, (c) 30 μm, (d) 35 μm, (e) 40 μm, (f) 45 μm, (g) 55 μm, (h) 60 μm, GT: ground truth z slice. (depth spacing=20 μm, GT=51, at 5338 μm away from CCD)

The GT is at 5338 μm away from CCD camera when a single particle with diameter equals 15 μm are firstly investigated, the depth spacing equals 20 μm. There are 50 reconstruction slices in each side of GT, hence overall 101 reconstruction slices, and the slice number of GT is 51. The maximum intensity curve and maximum gradient magnitude curve of the reconstruction stack utilizing Fresnel back propagation are depicted in Fig. 5(a) and Fig. 7(a), respectively. After $226^{th}$ iteration of the 3D HWNLD, the reconstructed images turn into Fig. 8(b). Most of the defocused images are diffused out; the reconstructed image at ground truth $z$ position is remained properly, including particle's shape and intensity. The maximum intensity of each reconstructed image is plotted in Fig. 9(a), wherein the maxima of the curve appears at the ground truth $z$ position, the maximum intensity of the other defocused images are almost 0. However, the defocused images in the neighboring slices of the ground truth $z$ position are not completely diffused out. The maximum gradient magnitude is plotted in Fig. 9(b) where the maxima of the curve appears at the ground truth $z$ position and the maximum gradient magnitude in neighboring slices are both much smaller than the maximum gradient magnitude in the GT. It implies that more iteration can fully diffuse out the neighboring defocused image. There are merely the data of central 101 slices depicted in Fig. 9, and the slice number of GT is 51.

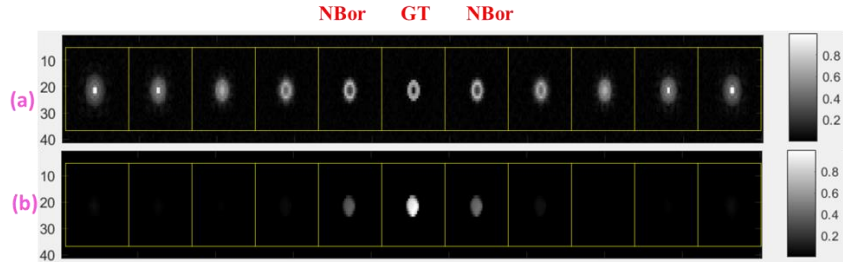

Fig. 8. (a) Central 11 reconstructed image after traditional Fresnel transform, (b) Central 11 reconstructed images after applying the proposed scheme, GT: ground truth $z$ slice, NBor, neighboring slice of GT.

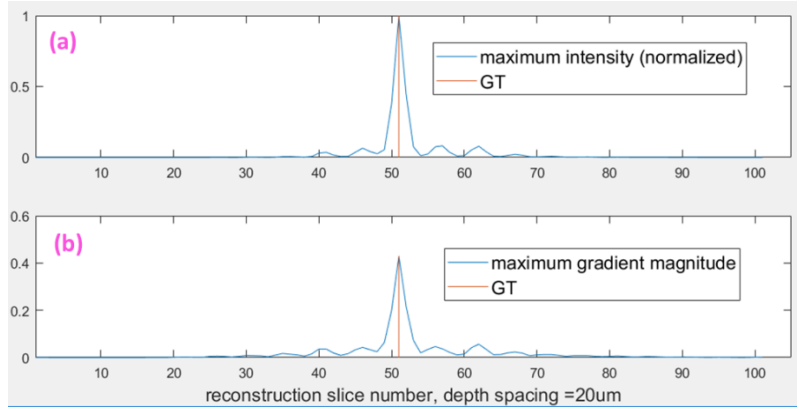

Fig. 9. (a) Maximum intensity, and (b) maximum gradient magnitude of each reconstruction slice in the reconstruction stack after applying the proposed scheme (101 reconstruction slices are shown, particle diameter = 15 μm, GT: ground truth z slice, GT=51, at 5338 μm away from CCD).

However, when a single particle with the diameter of 40 μm is investigated, in which the other parameters are as same as the investigation of the single particle with the diameter of 15 μm except that $\tau$ equals 2.2. After the $925^{th}$ iteration of 3D HWNLD, neither the maxima of the maximum intensity curve nor the maxima of the maximum gradient magnitude curve do not appear at the ground truth $z$ position, which are depicted in Fig. 10(a), and Fig. 10(b), respectively. The focused images converge at two positions which located at two sides of the ground truth $z$ position. Moreover, the *step-size* is decreased to 1/943720, which is significantly small so that the trend of the curve in Fig. 10(a) cannot change too much even though the program runs forever.

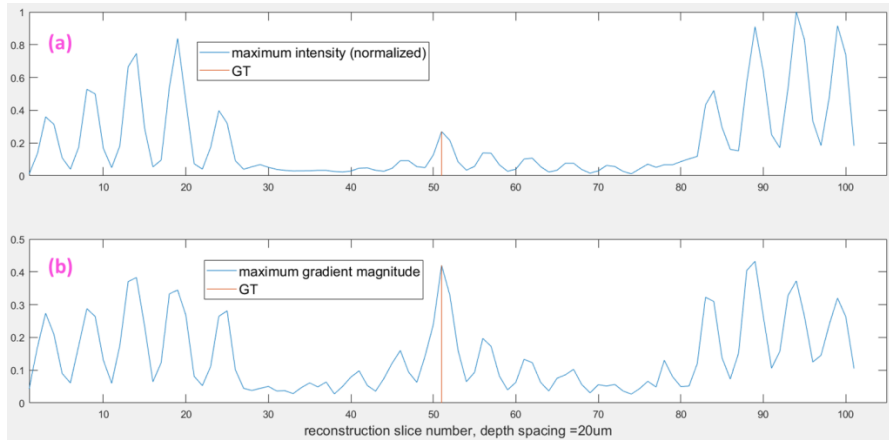

Fig. 10. (a)Maximum intensity, and (b)maximum gradient magnitude of the reconstruction stack after applying the proposed scheme (diameter = 40 μm, GT: ground truth z slice, GT=51, at 5338 μm away from CCD).

It is observed that the maxima of the maximum gradient magnitude curve does not always appear at the ground truth $z$ position when the axial reconstruction range is relatively long, and depth spacing is fine, such as the particle with the diameter of 25 μm, the reconstruction distance range is 2000 μm [4338 μm, 6338 μm], and reconstruction depth spacing equals 20 μm depicted in Fig. 7(b). However, the large magnitude gradient distribution that appears at the defocused image is too few to impact the distribution of the large gradient magnitude in the ground truth $z$ slice according to the red area upper the double-arrow line in Fig. 7, since the purpose of the proposed scheme is to remove the red area. Therefore, 3D HWNLD renders the focused image to converge at the ground truth $z$ position and renders the defocused images diffused out when the particle size is relatively small (not more than 35 μm in the case of this paper). It indicates that the sparsity distribution of the GT cannot be affected by the sparsity distribution of the corresponding defocused images with the reconstruction depth spacing of 20 μm when the particle is small enough.

### 3.3 Overlapping-particle case

When several particles are located at the same position in lateral plane but with different distance $z$, which indicates they are completely overlapping in the captured hologram, 3D HWNLD still can remove the defocused images and remain the focused image at the ground truth $z$ positions. Assume three particles with the diameters of 20 μm, 15 μm, and 25 μm are settled at 4538 μm, 5338 μm, and 6138 μm away from CCD camera, respectively. The hologram, whose mathematical expression is shown in Eq. (3), is shown in Fig. 6(b). After applying the proposed scheme, the central 11 reconstructed slices around the GT of each particle in the ranges of [4418 μm, 4618 μm], [5238 μm, 5438 μm], and [6038 μm, 6238 μm] with the depth spacing of 20 μm are depicted in Fig. 11, and the maximum intensity and maximum gradient magnitude of each reconstructed image in the reconstruction stack are plotted in Fig. 12(a) and (b), respectively. Even though a fairly long distance between two neighboring particles (800 μm at a minimum is tested) is required, it demonstrated that completely overlapping particles along $z$-axis can be resolved with small reconstruction depth spacing (20 μm) and defocused images can be completely removed. It is verified that if any particle with diameter less than 15 μm but not more than 35 μm, the defocused images will be diffused out as long as the distance between the two neighboring particles is much further. The simulation environment is the Windows 10 64-bit operating system, MATLAB R2017a, an Intel Core i7-8550U @1.80GHz CPU.

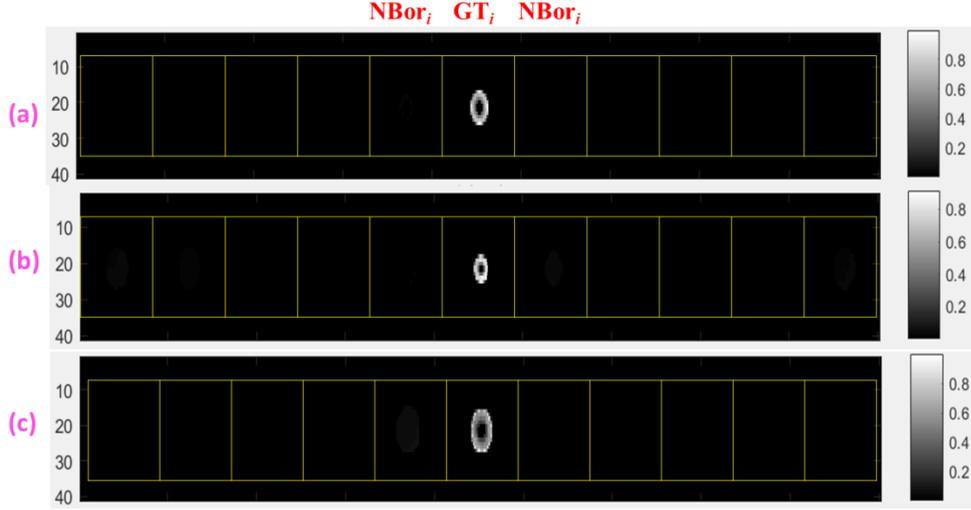

Fig. 11. The central 11 reconstructed images around the ground truth $z$ slice of (a) the first particle, diameter = 20 μm, (b) the second particle, diameter = 15 μm, and (c) the third particle, diameter = 25 μm, after applying the proposed scheme, $GT_i$: ground truth $z$ slice of each particle, $NBor_i$, neighboring slice of $GT_i$, $i$=1, 2, 3.

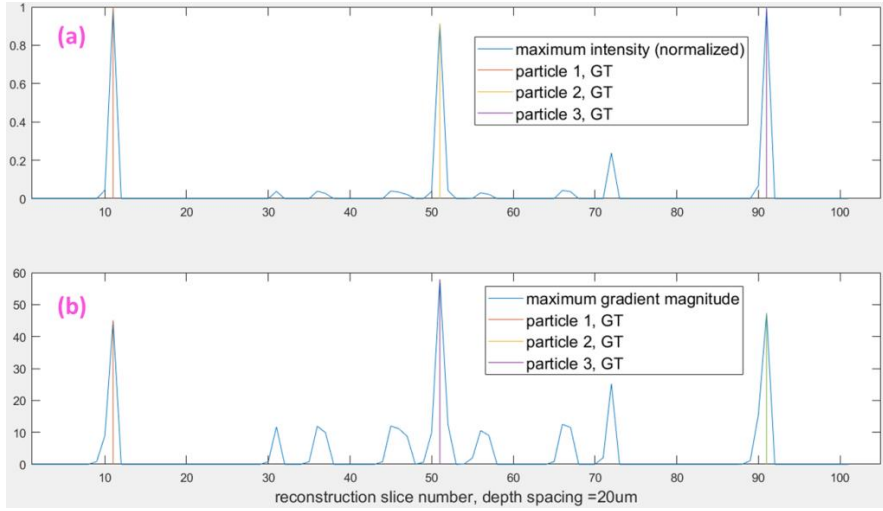

Fig. 12. (a) Maximum intensity, and (b) maximum gradient magnitude of each reconstructed image in the reconstruction stack of the three particles after applying the proposed scheme. (diameter of particle 1 = 20 μm, diameter of particle 2 = 15 μm, diameter of particle 3 = 25 μm, GT: ground truth z slice, $GT_1$=11, $GT_2$=51, $GT_3$=91, at 4538 μm, 5338 μm, and 6138 μm away from CCD, respectively)

## 4. Conclusions

We propose a 3D HWNLD method to implement the similar autofocusing function of multiple micro-objects and simultaneously remove the defocused images, which can also distinguish the locations of certain sized scattering particles that are overlapping along z-axis. It is applied to all of the reconstructed images after each back propagation. Despite the maxima of maximum gradient magnitude of each reconstruction slice in the original reconstruction stack does not appear at the ground truth $z$ position when the reconstruction range along $z$-axis is sufficiently long and the reconstruction depth spacing is sufficiently fine, the gradient magnitude of defocused image cannot affect the gradient distribution of the reconstructed image at ground truth *z* position when the particle is sufficiently small. Therefore the sparsity of the ground truth z slice cannot be affected so that the reconstructed image at ground truth *z* position is remained and the defocused images are diffused out after applying the proposed scheme. The numerical results also demonstrated that the proposed scheme can efficiently diffuse out the defocused images which are 20 μm away from the ground truth *z* position in spite of that several scattering particles with different diameters are completely overlapping along *z*-axis when the diameter not more than 35 μm and the hologram pixel pitch is 2 μm.

## Funding

This work was supported by Basic research program of Shenzhen (JCYJ20170412171744267) and National Basic Research Program of China (2015AA043302).

## Acknowledgements

The authors would like to thank Prof. George Barbastathis for his helpful suggestions.